\documentclass[conference]{IEEEtran}
\IEEEoverridecommandlockouts
\usepackage{booktabs} % For formal tables
\usepackage{amsmath}
\usepackage{comment}
\usepackage{bm}
\usepackage{graphicx}
\usepackage{balance}
\usepackage{array}
\usepackage{amssymb}% http://ctan.org/pkg/amssymb
\usepackage{pifont}% http://ctan.org/pkg/pifont
\usepackage{url}
\usepackage{microtype}
\usepackage{cleveref}
\usepackage{paralist}
\usepackage{enumitem}
\usepackage{tabularx}
\usepackage{tablefootnote}
\usepackage[flushleft]{threeparttable}
\usepackage{multirow}
\usepackage{balance}
\usepackage{subfig}
\usepackage[skip=\baselineskip]{caption}
\usepackage[space]{grffile}
\usepackage{xcolor}
\usepackage{cite}
\def\BibTeX{{\rm B\kern-.05em{\sc i\kern-.025em b}\kern-.08em
    T\kern-.1667em\lower.7ex\hbox{E}\kern-.125emX}}

\begin{document}

\title{Surfacing Estimation Uncertainty in the Decay Parameters of Hawkes Processes with Exponential Kernels}

\author{
\IEEEauthorblockN{Tiago Santos}
\IEEEauthorblockA{Graz University of Technology \\
Inffeldg. 16c, 8010 Graz, Austria \\
tsantos@iicm.edu}
\and
\IEEEauthorblockN{Florian Lemmerich}
\IEEEauthorblockA{RWTH Aachen University \\
Theaterplatz 14, 52062 Aachen, Germany \\
florian.lemmerich@cssh.rwth-aachen.de}
\and
\IEEEauthorblockN{Denis Helic}
\IEEEauthorblockA{Graz University of Technology \\
Inffeldg. 16c, 8010 Graz, Austria \\
dhelic@tugraz.at}
}

\maketitle

\begin{abstract}
As a tool for capturing irregular temporal dependencies (rather than resorting to binning temporal observations to construct time series), Hawkes processes with exponential decay have seen widespread adoption across many application domains, such as predicting the occurrence time of the next earthquake or stock market spike.
However, practical applications of Hawkes processes face a noteworthy challenge: There is substantial and often unquantified variance in decay parameter estimations, especially in the case of a small number of observations or when the dynamics behind the observed data suddenly change.
We empirically study the cause of these practical challenges and we develop an approach to surface and thereby mitigate them.
In particular, our inspections of the Hawkes process likelihood function uncover the properties of the uncertainty when fitting the decay parameter. We thus propose to explicitly capture this uncertainty within a Bayesian framework. 
With a series of experiments with synthetic and real-world data from domains such as ``classical'' earthquake modeling or the manifestation of collective emotions on Twitter, we demonstrate that our proposed approach helps to quantify uncertainty and thereby to understand and fit Hawkes processes in practice.
\end{abstract}


\section{Introduction}
As a method for modeling and predicting temporal event sequences (henceforth \textit{event streams}), Hawkes processes have seen broad application, ranging from estimating social dynamics in online communities~\cite{farajtabar2014shaping}, through measuring financial market movements~\cite{ait2015modeling} to modeling earthquakes~\cite{ogata1982application}.
Researchers and practitioners derive utility from Hawkes processes due to their flexibility in capturing history-dependent event streams. 
Hawkes processes model event streams via the conditional intensity function, the infinitesimal event rate given the event history. Events cause \textit{jumps} in the conditional intensity function, which \textit{decays} to a \textit{baseline} level following a pre-defined functional form, the so-called \textit{kernel}.
This kernel is often chosen as an exponential function. The reasons for this choice are manifold, as Hawkes processes with an exponential kernel are (i) efficient to simulate and estimate~\cite{liniger2009multivariate,farajtabar2015coevolve}, (ii) parsimonious, 
and (iii) realistic in practical applications~\cite{upadhyay2017uncovering,tabibian2019enhancing}. 

\begin{figure}[!t]
	\begin{center}
		\includegraphics[width=0.7\columnwidth]{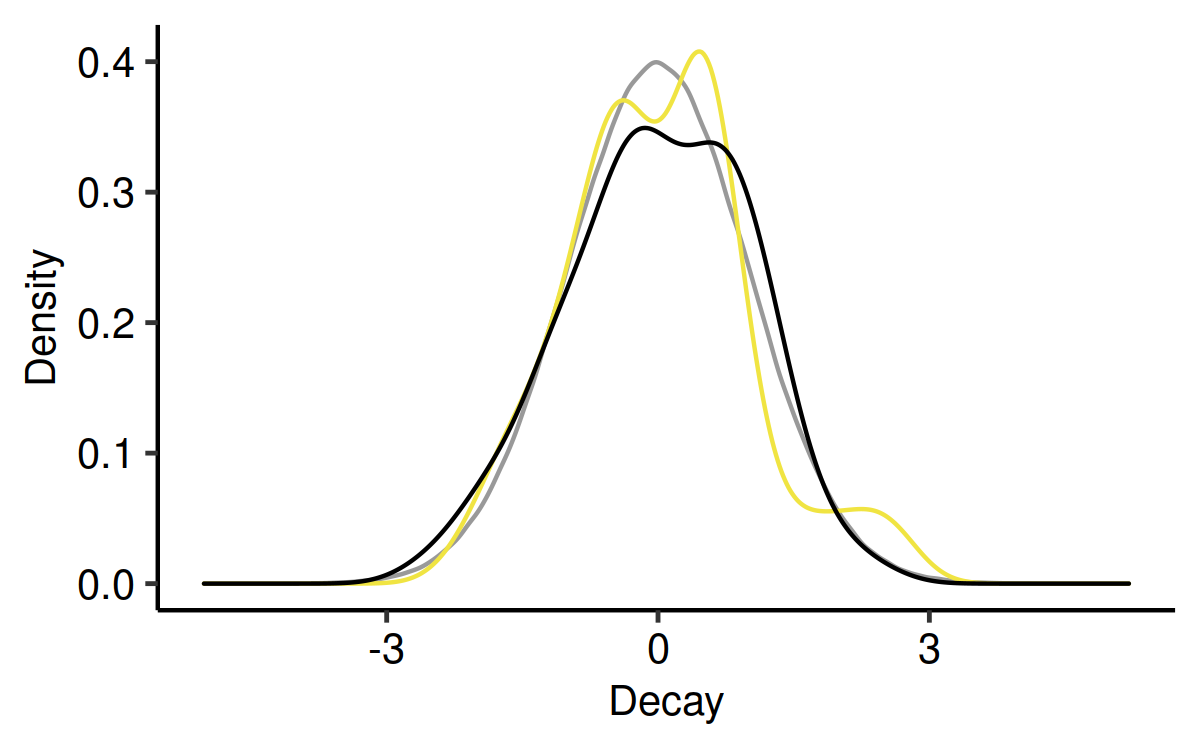}
		\caption{\label{fig:hawkes_decay_uncertainty_illustration}
			\textit{Remarkable Uncertainty in Fitted Decay Parameter Values.}
			We illustrate the normalized distribution of  decay values estimated with L-BFGS-B across $100$ realizations of the same Hawkes process (black line).
			The discrepancy between that distribution and the standard Gaussian distribution (gray line) suggests a unique uncertainty property in Hawkes process decay estimations.
			Fitting the decay parameter in Hawkes processes with breaks in stationarity results in yet other kinds of uncertainty properties (cf. yellow line).
			}
	\end{center}
\end{figure}

\noindent \textbf{Problem.} 
While baseline and jump parameters of Hawkes processes are typically derived via convex optimization, the estimation of decay parameters in exponential kernels remains an open issue.
Previous work simply assumed the decay parameters to be constants~\cite{farajtabar2015coevolve,2017arXiv170703003B,choudhari2018discovering}, 
cross-validated decay parameter values~\cite{farajtabar2014shaping,choi2015constructing,salehi2019learning}, or estimated them with a range of different optimization approaches~\cite{ozaki1979maximum,da2014hawkes,bacry2016mean,upadhyay2017uncovering,kurashima2018modeling,figueiredo2018fast,santos2019self}. Such estimation approaches result in point estimates that can be considered as sufficient for simulating and predicting event streams.
However, researchers frequently directly interpret Hawkes process parameters~\cite{ogata1982application,bacry2015hawkes,upadhyay2017uncovering,tabibian2017distilling}, e.g., to infer the directions of temporal dependency~\cite{junuthula2019block,trouleau2019learning,hatt2020early}.
Even though such applications rely on estimates of decay parameter values, the current state-of-the-art mostly neglects that, in practice, we derive these values only within a degree of certainty, in particular in a common case of a small number of realizations of a given process. This can lead to qualitatively inaccurate conclusions if researchers are unaware or unable to quantify the uncertainty of decay parameter estimate.
Moreover, up until now there are only initial studies~\cite{rizoiu2017expecting,santos2019self} on how exponential growth, exogenous shocks to a system, or changes in the underlying process mechanics compromise key stationarity assumptions and aggravate estimation errors.

In Figure~\ref{fig:hawkes_decay_uncertainty_illustration}, we illustrate this problem in relation to a commonly used nonlinear optimization approach for fitting the decay, L-BFGS-B~\cite{byrd1995limited}. We estimate the decay value in two cases: (i) a small number of realizations ($100$ synthetically generated realizations from the same Hawkes process), and (ii) change of process parameters due to a stationarity break ($50$ synthetically generated realizations from one Hawkes process and $50$ from a process with a different decay parameter). The black line in the Figure represents the normalized distribution of the fitted decay parameter for case (i). We observe that line deviates remarkably from the standard Gaussian depicted in gray, suggesting the need to account for the uncertainty surrounding Hawkes process decay parameter estimations. Repeating the same fitting procedure for case (ii), we observe, in the yellow line, yet other distributional properties of fitted decay values. This suggests a specific need to separately account for uncertainty in the decay parameter estimation in the presence of breaks in stationarity.

\noindent \textbf{This Work.} 
The contribution of this paper is two-fold: First, we focus on the potential uncertainty in the decay parameter of Hawkes processes and explore its sources and consequences.
In that regard, we uncover that the non-convex and noisy shape of the Hawkes process log-likelihood as a function of the decay is one cause of the variability in decay parameter estimates. 
Second, we propose to integrate the estimation of the decay parameter in a Bayesian approach. This allows for (i) quantifying uncertainty in the decay parameter, (ii) diagnosing estimation errors, and (iii) addressing breaks of the crucial stationarity assumption.
In our approach, we formulate and evaluate closed-form and intractable hypotheses on the value of the decay parameter. Specifically, we encode hypotheses for the decay as parameters of a prior distribution. Then, we consider estimations of the decay for single Hawkes process realizations as samples from a likelihood. These likelihood samples form the data that we combine with the prior to perform Bayesian inference of posterior decay values.

We show with synthetic data as well as in a broad range of real-word domains that our Bayesian inference procedure for fitting the decay parameter allows for quantifying uncertainty, diagnosing estimation errors and analyzing breaks in stationarity. 
In our first application, a study of earthquakes in Japanese regions~\cite{ogata1982application}, we uncover low uncertainty in certain geographical relationships. 
Second, we validate Settles and Meeder's~\cite{settles2016trainable} hypothesis that vocabulary learning effort on the Duolingo mobile app correlates to the estimated difficulty of the learned words~\cite{council2001common}. 
Finally, leveraging a dataset of Tweets before and after the Paris terror attacks of November $2015$, we measure the relation between a stationarity-breaking exogenous shock and collective effervescence~\cite{garcia2019collective}.

Overall, our work sheds light on fitting a widely used class of Hawkes processes, i.e., Hawkes processes with exponential kernels.
Better understanding these models and explicitly surfacing uncertainty in their fitted values facilitates their use by practitioners and researchers.
We expect the impact of our study to be broad\footnote{We make our code available at \url{https://github.com/tfts/hawkes_exp_bayes}.}, as
our results influence the application of a key analysis approach for studying time-dependent phenomena across most diverse domains such as social media, user behavior, or online communities.

\begin{figure}[!t]
	\begin{center}
		\includegraphics[width=0.7\columnwidth]{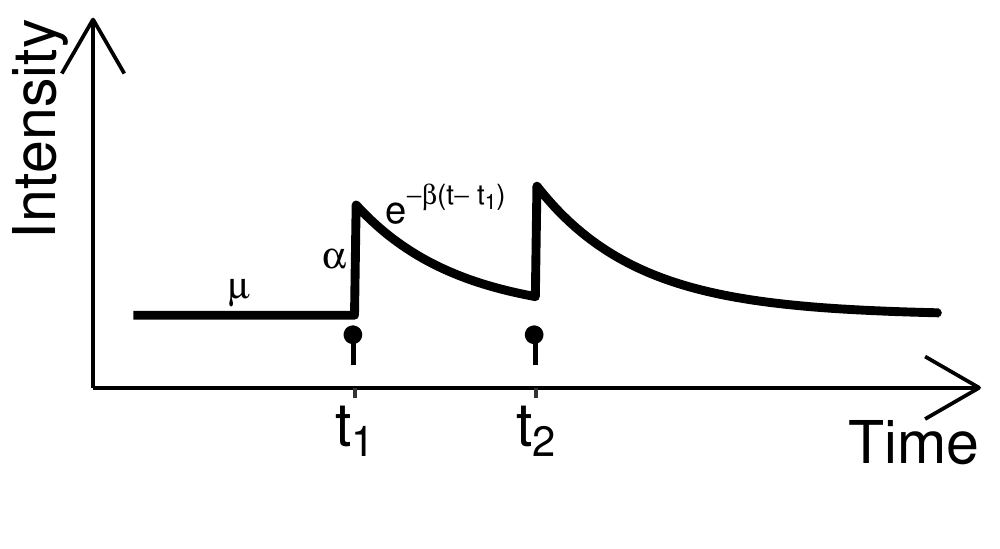}
		\caption{\label{fig:hawkes_intensity_illustration}
			\textit{Illustration of Hawkes Process Intensity $\lambda^*$.}
			This two event sample from a Hawkes process exemplifies how its intensity $\lambda^*$ changes over time. Starting with a minimal level of intensity $\mu$, $\lambda^*$ jumps by a constant $\alpha$ at each event time ${t_i}$, and then decays exponentially at rate $\beta$ over time.
		}
	\end{center}
\end{figure}

\section{Decay estimation in Hawkes Processes}
\label{sec:background}

In this section, we first summarize fundamentals of Hawkes process modelling before describing the problem of decay estimation.

\subsection{Hawkes Processes}

We briefly discuss temporal point processes, 
a set of mathematical models for discrete events randomly arriving over time. 
Temporal point processes capture the time of an upcoming event given the times of all previous events via the so-called \textit{conditional intensity function} (or simply \textit{intensity}) $\lambda^*(t)$. Mathematically, 

\begin{equation}
\lambda^*(t)dt=\text{P(event in }[t, t+dt)|\mathcal{H}_t),
\label{eq:general_intensity}
\end{equation}
where $\mathcal{H}_t$ represents the event history up to (but excluding) time $t$.
Dividing Equation~\ref{eq:general_intensity} by $dt$ (in the Leibniz notational sense), we see $\lambda^*(t)$ equals the conditional probability of an event in an interval of (infinitesimal) length $dt$ per such interval $dt$.
Hence, we interpret $\lambda^*(t)$ as a history-dependent \textit{event rate}.
Such temporal point processes are often termed doubly-stochastic, as events occur randomly over time, and the event model $\lambda^*(t)$ is a random process too.
We also note $\lambda^*(t)$ characterizes temporal point processes as counting processes $N(t)$ for the number of events up to time $t$.

Hawkes processes~\cite{hawkes1971spectra} assume $\lambda^*(t)$ follows a certain functional form. Specifically, given a Hawkes process \textit{realization}, i.e., a set of $n$ events occurring at times $t_i \in \mathbb{R}^+$, the conditional intensity  of a Hawkes process is
\begin{equation}
\lambda^*(t) = \mu + \alpha\sum_{t_i \in \mathcal{H}(t)} \kappa_{\beta}(t - t_i),
\label{eq:hawkes_intensity}
\end{equation}
where $\mu \in \mathbb{R}^+$ is the \textit{baseline} intensity 
and $\alpha \in \mathbb{R}^+$ the \textit{self-excitation}, i.e., magnitude of an increase in  $\lambda^*(t)$ at each event time $t_i$. Immediately after each $t_i$, the intensity decreases according to the kernel $\kappa_{\beta}$. A common choice~\cite{zhou2013learning,farajtabar2014shaping,farajtabar2015coevolve,upadhyay2017uncovering,santos2019self,trouleau2019learning} for the kernel is an exponential function parametrized by the \textit{decay} rate $\beta$, i.e., $\kappa_{\beta}(t) = e^{-\beta t}, \beta \in \mathbb{R}^+$. Plugging this kernel in Equation~\ref{eq:hawkes_intensity} we obtain
\begin{equation}
\lambda^*(t) = \mu + \alpha \sum_{t_i \in \mathcal{H}(t)} e^{-\beta (t-t_i)},
\label{eq:hawkes_exp_kernel}
\end{equation}
which we illustrate in Figure~\ref{fig:hawkes_intensity_illustration}.

Multivariate Hawkes processes with an exponential kernel generalize univariate ones by introducing parameters for self-excitation and for the decay per dimension. Beyond self-excitation, they also capture \textit{cross-excitation}, the intensity jump an event in one dimension causes in another. Formally, the intensity of dimension $p$ of an $M$-variate Hawkes process is

\begin{eqnarray}
\lambda^{*p}(t)&=&\mu_p+\sum_{q=1}^M\sum_{t_i^q<t} \alpha_{pq} e^{-\beta_{pq} (t-t_i^q)}.
\label{eq:hawkes_intensity_m}
\end{eqnarray}
Notice that we index each dimension's intensity function with $\lambda^{*p}$, and its baseline with $\mu_p$. This generalization also includes an excitation matrix with self-excitation and cross-excitation entries $\alpha_{pp}$ and respectively $\alpha_{pq}$, as well as, analogously, a matrix of decay rates $\beta_{pq}$. Note that $\alpha_{pq}$ captures the increase in intensity in dimension $p$ following an event in dimension $q$.
In matrix notation, we write $\bm{\mu} \in \mathbb{R}^M$, $\bm{\alpha} \in  \mathbb{R}^{M\times M}$ and $\bm{\beta} \in  \mathbb{R}^{M\times M}$.
We point to the work by Linniger~\cite{liniger2009multivariate} for more on multivariate Hawkes process theory.

We now introduce the notions of \textit{stationarity} and \textit{causality} in the multi-dimensional Hawkes process context. 
Stationarity implies translation-invariance in the Hawkes process distribution, and, in particular, it also implies that the intensity does not increase indefinitely over time and therefore stays within bounds. More formally, a Hawkes process with an exponential kernel is stationary if the spectral radius $\rho$, i.e., the largest eigenvalue of the $L^1$-norm of $\bm{\alpha} / \bm{\beta}$, satisfies $\rho < 1$. 
Note that assessing stationarity of a one-dimensional Hawkes process with an exponential kernel reduces to evaluating $\alpha / \beta < 1$. 

Recent work~\cite{etesami2016learning} established a connection between Granger causality and the excitation matrix: In the particular case of our exponential kernel, dimension $q$ Granger-causes dimension $p$ if and only if $\alpha_{pq}>0$. Beyond this result, we interpret the magnitude of excitations $\alpha_{pq}$ as the strength and direction of temporal dependency between dimensions $p$ and $q$: For example, we say dimension $q$ influences dimension $p$ more strongly if and only if $\alpha_{pq} > \alpha_{qp}$.

Finally, we define the Hawkes process likelihood function. We work with the log-likelihood function due to its mathematical manipulability and to avoid computational underflows. The equation for the one-dimensional log-likelihood of the Hawkes process with an exponential kernel is as follows~\cite{ozaki1979maximum}:
\begin{eqnarray}
log\,L(\{t_i\}_{i=1}^n) = -\mu t_n - \frac{\alpha}{\beta} \sum_{i=1}^n{(1-e^{-\beta (t-t_i)})} \nonumber\\+ \sum_{i=1}^n{log(\mu+\alpha A(i))},
\label{eq:hawkes_loglik}
\end{eqnarray}
\[\text{ with A(1) = 0 and, for } i>1, \, A(i) = \sum_{t_j < t_i}{e^{-\beta (t_i-t_j)}}.\]
Ozaki~\cite{ozaki1979maximum} also proposes a (computationally less intensive) recursive formulation for Eq.~\ref{eq:hawkes_loglik}. We refer to Daley and Vere-Jones~\cite{daley2003introduction} for a general formulation of the log-likelihood of multivariate temporal point processes. 
Henceforth, as we focus on Hawkes processes with exponential kernels, we refer to them simply as Hawkes processes.

\begin{figure*}[t!]
	\begin{center}
		\subfloat[Large Range]{
			\includegraphics[width=0.31\textwidth]{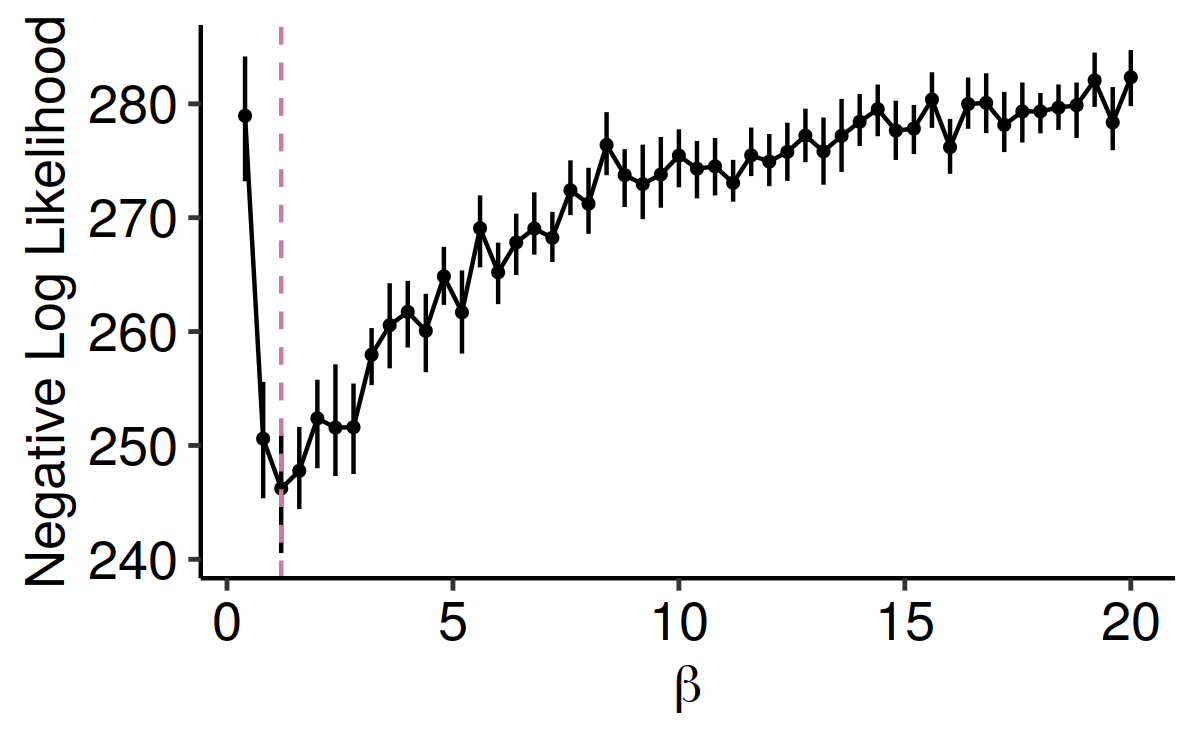}
			\label{fig:loglik_far}
		}
		\subfloat[Medium Range]{
			\includegraphics[width=0.31\textwidth]{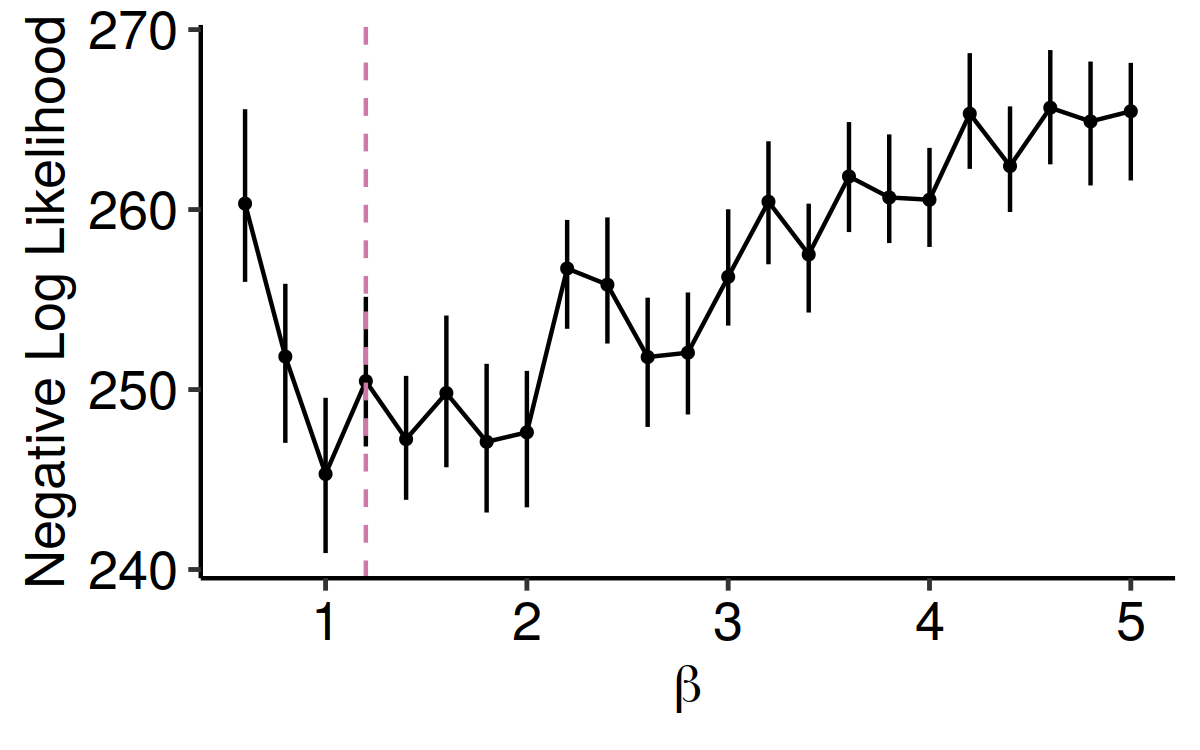}
			\label{fig:loglik_broad}
		}
		\subfloat[Small Range]{
			\includegraphics[width=0.31\textwidth]{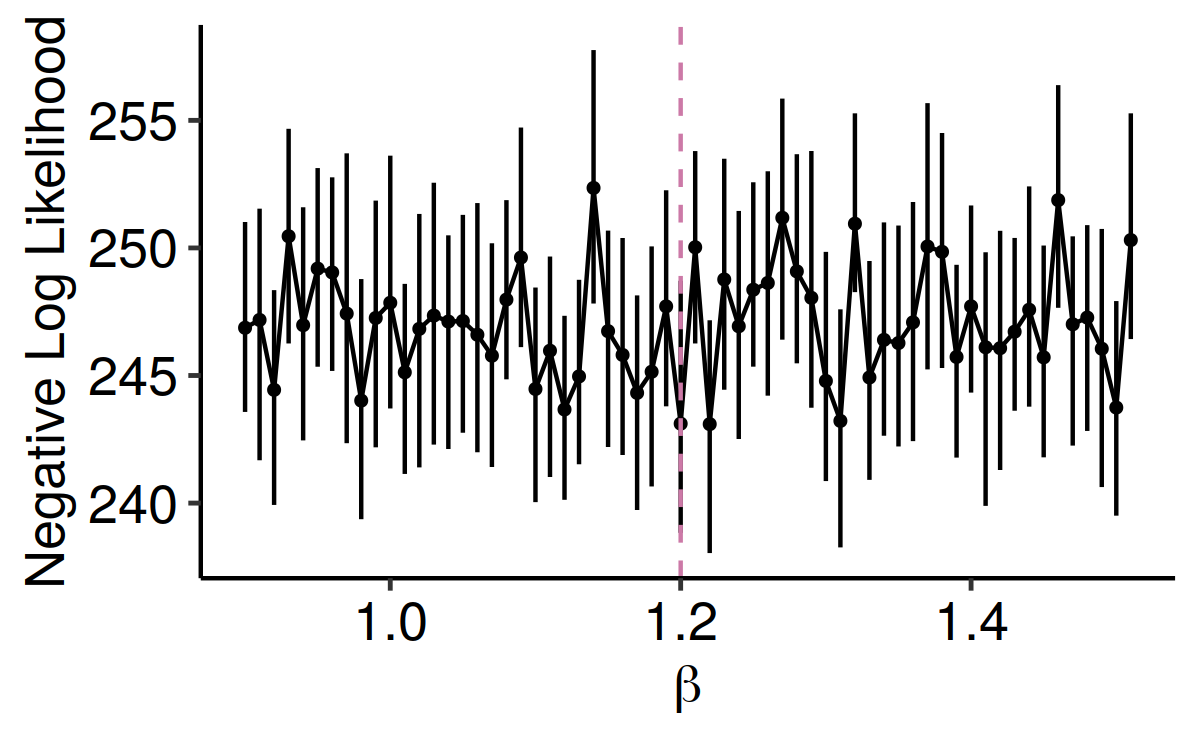}
			\label{fig:loglik_close}
		}
		\caption{\label{fig:loglik}
			\textit{The Negative Log-Likelihood of Hawkes Processes as a Function of $\beta$ Is Non-Convex And Noisy.} 
			We simulate three separate sets of $100$ realizations from a one-dimensional Hawkes process with $\beta^*=1.2$ (marked with a pink dashed line), and we evaluate the process' log-likelihood with one set of realizations per each of three ranges of $\beta$ around $\beta^*$: a large (cf. Fig.~\ref{fig:loglik_far}), a medium (cf. Fig.~\ref{fig:loglik_broad}) and a small (cf. Fig.~\ref{fig:loglik_close}) range. Error bars indicate $95\%$ confidence intervals. 
			In Fig.~\ref{fig:loglik_far}, it is apparent the negative log-likelihood is neither concave nor convex in $\beta$. Zooming in around $\beta^*$ reveals a wide and noisy basin and, in Fig.~\ref{fig:loglik_broad}, the negative log-likelihood is not minimal at $\beta^*$. Both observations explain difficulties in optimizing the log-likelihood for the decay parameter. This finding is robust across Hawkes process parameter configurations satisfying the stationarity constraint.
		}
	\end{center}
\end{figure*}

\subsection{Decay Estimation}
\label{sec:fitting}
To learn from streams of events, practical applications start by fitting Hawkes processes, i.e., optimizing the log-likelihood given in Equation~\ref{eq:hawkes_loglik} to a set of event timestamps. 
Practitioners then inspect fitted parameters to understand inherent temporal dependencies, and perform downstream tasks such as prediction via simulation of the fitted processes~\cite{kurashima2018modeling,santos2019self}. 
As inferring and interpreting (all) fitted Hawkes process parameters is crucial in many real-world applications~\cite{ogata1982application,bacry2015hawkes,tabibian2017distilling,upadhyay2017uncovering,santos2019self,junuthula2019block,hatt2020early}, we turn our attention to the challenges in fitting Hawkes process parameters, and especially, in estimating the decay parameter in the exponential kernel.
Previous research has shown that the baseline $\mu$ and excitation jump $\alpha$ can be efficiently computed since the log-likelihood is amenable for convex optimization of these parameters \cite{bacry2015hawkes, farajtabar2015coevolve}.
However,
that does not hold for the decay $\beta$, neither in the univariate nor in the multivariate case.

Previous work suggested a wide range of methods to address the decay estimation problem with approaches that provide point estimates. 
These approaches include setting $\beta$ to a given constant value~\cite{farajtabar2015coevolve,2017arXiv170703003B,choudhari2018discovering}, cross-validation over a range of values~\cite{farajtabar2014shaping,choi2015constructing,salehi2019learning}, or the application of a general optimization method. Those methods comprise non-linear optimization~\cite{ozaki1979maximum,da2014hawkes},  
Bayesian hyperparameter optimization~\cite{figueiredo2018fast,santos2019self}, expectation-maximization~\cite{kurashima2018modeling,turkmen2019hawkeslib} or visual inspection of the log-likelihood function~\cite{bacry2016mean,upadhyay2017uncovering}. 
While these methods suffice for obtaining a point estimate with comparable errors, none of them explicitly addresses  uncertainty quantification.

We argue that aforementioned problems in estimating decay parameters (cf. Figure~\ref{fig:hawkes_decay_uncertainty_illustration}) are caused by a noisy, non-convex log-likelihood in $\beta$. 
We illustrate these properties of the log-likelihood with the following exemplary experiment. We consider a univariate Hawkes process with $\mu=0.1, \alpha=0.5$ and $\beta^*=1.2$ (the same as in Fig.~\ref{fig:hawkes_decay_uncertainty_illustration}), and we then compute the negative log-likelihood for different values of $\beta$.
We generate three sets of $100$ realizations from that Hawkes process.
In Figure~\ref{fig:loglik}, we evaluate the negative log-likelihood with one set of realizations per each of three ranges of $\beta$ around $\beta^*$, namely a large (cf. Fig.~\ref{fig:loglik_far}), a medium (cf. Fig.~\ref{fig:loglik_broad}) and a small (cf. Fig.~\ref{fig:loglik_close}) range.
In the large range, it appears there is a convex basin around $\beta^*$ (which we annotate with a pink dashed line), but this function's shape shifts to a concave curve with increasing decay values. The function then converges on the right, as $\lim_{\beta \to +\infty} log\, L=-\mu t_n +nlog(\mu)$.
Inspecting the seemingly ``convex'' region more closely uncovers a wide and noisy basin around $\beta^*$, where $\beta^*$ does not always feature minimal negative log-likelihood (cf. Fig.~\ref{fig:loglik_broad}).
This explains difficulties in obtaining precise estimations regardless of the optimization strategy.
We repeated this experiment with a wide range of parameter values corresponding to stationary Hawkes processes. We confirm that these observations are robust to all such alternative configurations.

\section{Bayesian Decay Estimation}\label{sec:approach}
In this section, we present a novel Bayesian approach for the decay parameter estimation in Hawkes process. We begin with the overall goals and the intuition behind our approach before formally introducing the approach and its operationalization. We also provide a short reflection on a frequentist alternative.

\noindent \textbf{Goals.}
The aim of our approach is threefold. Firstly, we try to explicitly quantify $\beta$ estimation uncertainty and the magnitude of potential estimation errors across $\beta$ fitting approaches (i.e., point estimates). 
Secondly, considering previous use cases~\cite{ogata1982application} that involve encoding and validating hypotheses on the decay value, we aim to systematically diagnose such hypotheses.
Thirdly, current methods return only decay values which fulfill the stationarity constraint $\rho < 1$. However, previous work~\cite{rizoiu2017expecting,santos2019self} studied applications with non-stationary changes such as exponential growth and exogenous shocks. Hence, we see an opportunity for an extension of current decay fitting methods to address potential violations of the stationarity assumption.

\noindent \textbf{Intuition.}
To that end, we propose a parsimonious Bayesian inference procedure for encoding and validating hypotheses on likely values for $\beta$. 
In our Bayesian approach, we sequentially consider a series of univariate Hawkes process realizations one by one. With each subsequent realization, we fit $\beta$ with a given optimization method and obtain an ever-increasing set of estimated decay values that we denote as $\{\hat{\beta}\}_{k=1}^K$~\footnote{Note that the set $\{\hat{\beta}\}$ does not contain independent observations (as e.g. the realizations used to obtain $\{\hat{\beta}\}_{k=1}^2$ are also in $\{\hat{\beta}\}_{k=1}^3$). 
Anticipating that practitioners use as much data as available, we consider $\{\hat{\beta}\}$ as previously defined.  However, repeating all our experiments using only iid $\hat{\beta}_k$ (i.e., each fitted only on a single realization), we obtain similar though noisier results.}.
After this collection of $\hat{\beta}$, we apply Bayes' theorem to make inferences about the true $\beta$.

A key difference between our approach and typical applications of Bayesian inference can intuitively be described as follows:
The classical Bayesian inference setup typically places a prior distribution for an unknown parameter of interest in a probability distribution which captures the likelihood of given data. Then, applying Bayes' theorem allows for inferring likely values of the unknown parameter given the data.
In our approach, we assume the unknown parameter, i.e., the decay parameter, is also the data, which consists of the aforementioned sequence of decay parameter estimations. This setup enables the freedom to choose between computing posterior (i.e., in parameter space) or posterior predictive (i.e., in data space) distributions to learn about the decay. We find that this flexibility is useful and can improve performance in applications.

\noindent\textbf{Formalization.} 
Given data $\{\hat{\beta}\}_{k=1}^K$ and a model $\mathcal{M}_H$ parametrized by hypothesis $H$ for the parameter of interest $\beta$, we propose computing the Bayesian posterior 
\begin{equation}
P(\beta|\{\hat{\beta}\}_{k=1}^K, \mathcal{M}_H) \propto P(\{\hat{\beta}\}_{k=1}^K|\beta, \mathcal{M}_H) P(\beta|\mathcal{M}_H),
\label{eq:bayes}
\end{equation}
where $P(\{\hat{\beta}\}_{k=1}^K|\beta, \mathcal{M}_H)$ is the likelihood and $P(\beta|\mathcal{M}_H)$ the prior encoding a hypothesis. Again, note the accordance between the model parameter and the data in our Bayesian approach: $\beta$ is the model parameter and, at the same time, our data $\{\hat{\beta}\}_{k=1}^K$ contains the estimates of that same parameter. Having derived or estimated a posterior distribution density, we have multiple inference possibilities for $\beta$. In particular, we can, as introduced above, (i) obtain a point estimate for $\beta$ directly from the posterior (e.g., mean, median, or maximum), or (ii) compute a new estimate $\tilde{\beta}$ or a statistic (e.g., mean) from the posterior predictive distribution, which is given by (omitting for simplicity $\mathcal{M}_H$):

\begin{equation}
P(\tilde{\beta}|\{\hat{\beta}\}_{k=1}^K) = \int P(\tilde{\beta}|\beta, \{\hat{\beta}\}_{k=1}^K)P(\beta|\{\hat{\beta}\}_{k=1}^K)d\beta.
\label{eq:bayes_pred}
\end{equation}
We can also derive credible intervals ---roughly Bayesian counterparts of the frequentist confidence interval--- for such statistics.

\noindent\textbf{Operationalization.}
Thus, this setup leaves us with the task of deciding on appropriate distributions for the prior and the likelihood, as well as mapping hypotheses to specific parametrizations of the prior.
While we are, in general, free to choose from the broad spectrum of available techniques for Bayesian inference, we propose to assume the likelihood $P(\{\hat{\beta}\}_{k=1}^K|\beta) \sim Exponential(\lambda)$ and a conjugate prior, namely $P(\beta) \sim Gamma(a_0, b_0)$. 
These assumptions lead to the following practical advantages: (i) there is a closed form solution to the posterior predictive density mean, 
$\beta' = E[\tilde{\beta}|\{\hat{\beta}\}_{k=1}^K] = (b_0 + \sum_{k=1}^K{\hat{\beta}_k}) / (a_0-1)$, and 
(ii) users of this approach only have to encode hypotheses on $\beta$ as the prior's $b_0$ parameter, as they can set $a_0$ simply to $K$.
We note that assuming other distributions of the exponential family such as Pareto for the log-likelihood and the conjugate prior is also a valid choice, for which we experimentally obtain equivalent results. 
Note that our choice of a Bayesian setup also allows for complex inference setups, which require the application of e.g. Markov chain Monte Carlo (MCMC). 
Also, in our approach, we restrict Bayesian inference to the decay and estimate confidence intervals for other parameters via (frequentist) bootstrap.
This enables the estimation of other parameters via more efficient convex optimization routines.
We also remark that our approach is agnostic to the choice of decay estimation method, and, in our experiments, we illustrate its application with several of the methods we mentioned in Section~\ref{sec:fitting}.

So far, we focused the presentation of our approach on univariate Hawkes processes. To generalize to the multivariate case, we set $\beta_{pq}=\beta \text{  } \forall_{p,q}$, a common simplification of the decay estimation problem~\cite{farajtabar2014shaping,tabibian2017distilling,santos2019self,salehi2019learning}, and we then proceed as previously.
Alternatively, optimization of $\beta_{pq}$ is also possible, but with a quadratic computational complexity in the number of dimensions.

\noindent\textbf{A Frequentist Alternative.}
Finally, we reflect on a bootstrap-based alternative formulation of our Bayesian approach. Repeatedly resampling $\{\hat{\beta}\}_{k=1}^K$ with replacement yields the bootstrap distribution of decay estimates. Testing hypotheses on the decay value then amounts to comparing a statistic of the bootstrap distribution (such as a mean or a percentile)\footnote{We recommend using empirical bootstrap statistics to correct for sample bias.} with the hypothesized value. While it is also possible to extend this bootstrap-based procedure to more complex, intractable inferential setups, the computational cost may surpass that of Bayesian inference (as we discuss in the context of an experiment). In our setting of sequential collection of realizations, Bayesian inference becomes a natural choice to update beliefs about true decay values (cf. Chapter 3.4 of Efron and Hastie~\cite{efron2016computer}), especially in the case of a small number of realizations. Further, as the bootstrap may be interpreted as a nonparametric, noninformative posterior approximation (cf. Chapter 8.4 of Hastie et al.~\cite{hastie2009elements}), Bayesian inference with informative priors may, again, lead to lower computational costs in the presence of few observations.

\section{Experiments}
\label{sec:experiments}
We illustrate experimentally how our approach allows for quantifying the uncertainty in the decay and other Hawkes process parameter values, diagnosing mis-estimation and addressing breaks in stationarity.
For each of those three goals, we illustrate how our approach achieves the goal (i) with a synthetic dataset and (ii) within a real-world application. Hence, we demonstrate that, besides achieving the previously mentioned goals, our approach is broadly applicable across practical scenarios.

\subsection{Quantifying the Uncertainty}
We begin by addressing the problem of quantifying the uncertainty of fitted decay values $\hat{\beta}$, as well as potential consequences of mis-estimation on other Hawkes process parameters.
One prominent application of multi-dimensional Hawkes processes consists in the estimation of directions of temporal dependency, e.g., when studying influence in online communities~\cite{santos2019self,junuthula2019block}, or in approximating complex geographical~\cite{ogata1982application} or cortical~\cite{trouleau2019learning} relationships.
Recall that inferring such relations between a pair of Hawkes process dimensions may be framed as a problem of estimating which cross-excitation between the two dimensions is higher.
We demonstrate how our Bayesian approach helps in quantifying the uncertainty in such inferred relationships and how to estimate the impact of potential errors with synthetic and real-world data.

\noindent\textbf{Synthetic Data.}
We consider a two dimensional Hawkes process with parameters $\bm{\mu} = \left(\begin{smallmatrix}0.1\\0.5\end{smallmatrix}\right)$, 
$\bm{\alpha} = \left(\begin{smallmatrix}0.1&\alpha_{12}\\\alpha_{21}&0.2\end{smallmatrix}\right)$, and 
$\beta_{pq}=\beta \text{  } \forall_{1 \leq p,q \leq 2}$. We assume $\beta=1.2$.
For the cross-excitation parameters, we set $\alpha_{21} = 0.7$ and, successively, $\alpha_{12} = \alpha_{21} * c$ for a range of $10$ linearly spaced values of $c\in[0.75,1.25]$. This implies that each configuration encodes a different direction and strength of influence, where dimension $2$ more strongly influences dimension $1$ for $c < 1$, and vice-versa for $c > 1$. 
For each such configuration, we simulate $K=100$ realizations with a stopping time of $T=1000$. 
As the decay estimation approach we apply  a non-linear optimization routine L-BFGS-B~\cite{byrd1995limited} after the arrival of each such realization to obtain $\{\hat{\beta}\}_{k=1}^{100}$ per configuration.
Using our closed-form Bayesian inference approach, we hypothesize that $\beta$ equals $1.5$ by setting $b_0=1.5$ in the previously described Gamma prior. 
We then perform the aforementioned Bayesian inference on each set of $\{\hat{\beta}\}_{k=1}^{100}$ to derive $\beta'_{0.025}$ and $\beta'_{0.975}$, the lower and upper bounds of the $95\%$ credible interval of the posterior predictive density.
For $100$ linearly spaced values of the decay in $[\beta'_{0.025}, \beta'_{0.975}]$, we fit the remaining Hawkes process parameters (i.e. baseline and excitations) and check the accuracy of the inferred influence direction between dimensions.
The accuracy captures how many of the $100$ decay values lead to correct recovery of the relation between $\alpha_{12}$ and $\alpha_{21}$. We bootstrap $95\%$ confidence intervals\footnote{We report bootstrap statistics here to refrain from unnecessarily extending our Bayesian approach to the other Hawkes process parameters, which, as previously mentioned, should be estimated with convex optimization routines.} for the accuracy directly from that distribution of $100$ decay values.
This procedure provides 
\begin{inparaenum}[(i)]
	\item an estimate of the uncertainty of the fitted decay value via the $95\%$ credible interval,
	\item an estimation of the robustness of the temporal dependency between dimensions $1$ and $2$, and
	\item empirical evidence for the consequences of fitting Hakwes processes with misaligned $\beta$.
\end{inparaenum} 
Beyond the settings presented here, we also experimented  with slightly different parametrizations of the Hawkes process and alternative decay fitting approaches such as the Bayesian hyperparameter optimization Hyperopt~\cite{bergstra2011algorithms}. In all such settings, we obtained qualitatively the same results. 

\begin{figure}[!t]
	\begin{center}
		\includegraphics[width=0.7\columnwidth]{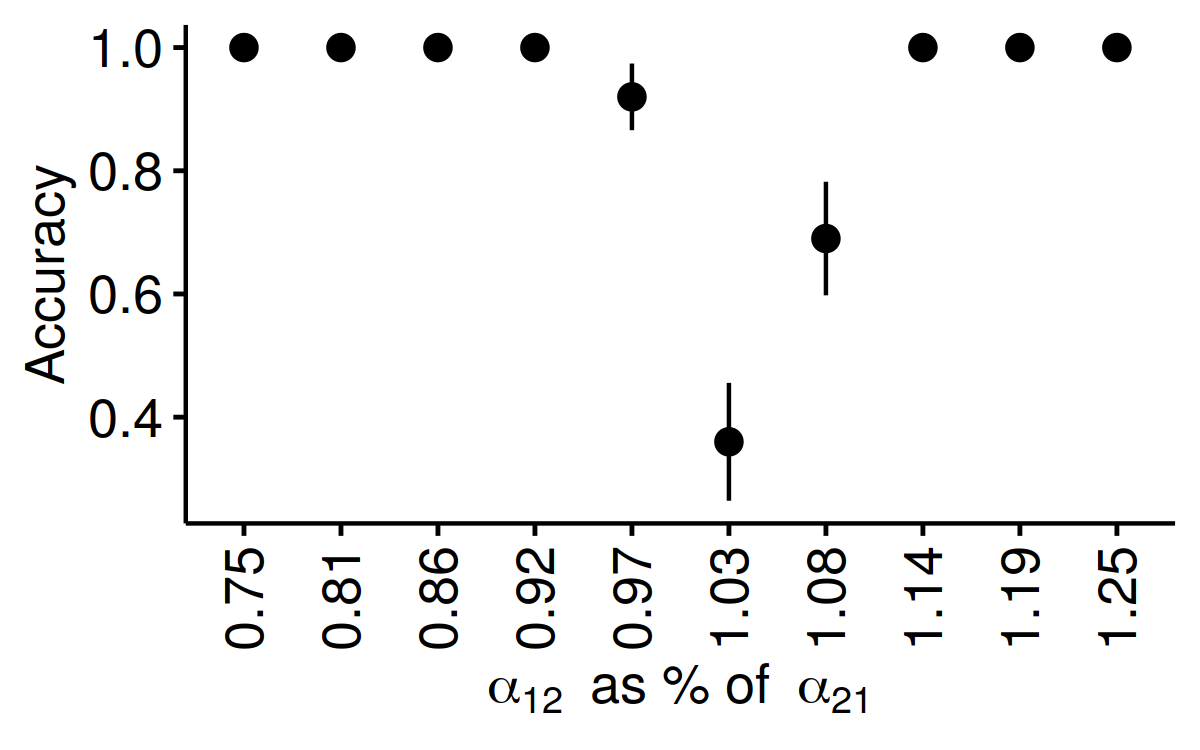}
		\caption{\label{fig:syn_uncertainty}
			\textit{Quantifying Uncertainty When Inferring Directions of Influence.}
			We first fit $\beta$ on realizations from two-dimensional Hawkes processes with cross-excitation $\alpha_{12}$ varying from $75\%$ to $125\%$ of $\alpha_{21}$.
			We apply closed-form Bayesian inference to estimate the uncertainty in fitted decay as the posterior predictive $95\%$ credible interval. 
			For a set of decay values in that interval, we estimate the other parameters, and measure the accuracy in recovering the encoded influence direction between dimensions $1$ and $2$.
			The accuracy is low and features larger error bars when $\alpha_{12}$ is close to $\alpha_{21}$, where many decay values in the $95\%$ credible intervals lead to wrong estimations of the influence direction.
		}
	\end{center}
\end{figure}

Figure~\ref{fig:syn_uncertainty} summarizes the outcome of this experiment. 
As expected, we observe lower accuracies in inferring the direction of influence between dimension $1$ and dimension $2$ for $\alpha_{12}$ close to $\alpha_{21}$.
This implies that many decay values in the $95\%$ credible interval of such configurations lead to mis-estimations of the direction of influence, and the large error bars reflect this as well.
Overall, we believe practitioners may leverage this approach to quantify the uncertainty of the decay estimation and consequently of the inferred directions of influence as well.

\noindent\textbf{Earthquakes and Aftershocks.}
We illustrate the outlined uncertainty quantification procedure with a dataset of earthquakes in the Japanese regions of Hida and Kwanto, as originally studied by Ogata et. al~\cite{ogata1982application}.
We consider the data listed in Table $6$ of that manuscript, i.e., a dataset of $77$ earthquakes from $1924$ to $1974$.
We employ the decay value listed in Table $5$ of that manuscript as the prior's parameter in our closed-form Bayesian approach\footnote{We choose that prior for demonstration purposes, as that decay value was estimated with a different Hawkes process from the one we study in this work.}. We assume a two-dimensional Hawkes process with a single $\beta$ value, where the dimensions represent earthquakes in the Japanese regions of Hida and Kwanto. 
As pre-processing, we split earthquake occurrences into $K=4$ equally sized segments which we treat as process realizations, and we convert the event timescale to decades. 

We replicate the seismological relationship that earthquakes in the Japanese Hida region precede those in Kwanto, 
as we obtain $\alpha_{\text{Kwanto}\,\text{Hida}} > \alpha_{\text{Hida}\,\text{Kwanto}} = 0$. 
Our Bayesian estimation procedure yields the posterior predictive density mean of $\beta' \approx 31.71$, which corresponds to an intensity half-life of about $log(2) / 31.71 \approx 0.02$ decades.
The inferred relationship between Hida and Kwanto is present for all but one value of the $95\%$ credible interval for $\beta'$, the lower extremity $\beta'_{0.025}$. 
These results underscore the low uncertainty of the inferred direction of influence.

\subsection{Mis-Estimation \& Misaligned Hypotheses}
In this section, we demonstrate that our Bayesian approach facilitates diagnosing (inevitable) estimation errors and misaligned hypotheses as over- or under-estimates, as well as the magnitude of that error.
Hence, we address a need, which previous work~\cite{ogata1982application,bacry2016mean,upadhyay2017uncovering} implies, to encode, validate and diagnose estimations and hypotheses on the decay parameter value.
Again, we illustrate how our approach meets that need with synthetic and real-world data.

\noindent\textbf{Synthetic Data.} 
\label{subsubsec:closed_bayes}
We consider a univariate Hawkes process with parameters $\mu=1.2, \alpha=0.6$ and $\beta=0.8$ (comparable choices of parameters lead to the same qualitative results).
Using this parametrization, we generate $K=100$ realizations with $100$ events each and we estimate decay after the arrival of each individual realization to obtain $\{\hat{\beta}\}_{k=1}^{100}$ per approach. To illustrate that our Bayesian approach is agnostic to the choice of decay estimation method, we apply the decay estimation approaches L-BFGS-B~\cite{byrd1995limited}, Grid Search (across $10$ evenly distributed values on a log scale in $[-1, 2]$, similarly to Salehi et al.~\cite{salehi2019learning}) and Hyperopt~\cite{bergstra2011algorithms}. 
Using our closed-form Bayesian inference approach, we leverage a Gamma prior with $b_0=1$, an estimate slightly larger than true $\beta$, to illustrate our approach.
We then perform Bayesian inference on each set of $\{\hat{\beta}\}_{k=1}^{100}$ and compare the RMSE (root mean squared error) between the resulting $\beta'$ estimates and $\beta$.
We repeat this whole process $100$ times to derive uncertainty per fitting method via the Bayesian bootstrap distribution.

We observe that L-BFGS-B and Hyperopt return the decay estimates with lowest RMSE, while Grid Search performs remarkably worse.
Aiming to reduce RMSE across all approaches, we inspect the estimates more closely.
Looking into $\hat{\beta}_k$ reveals that they are consistently below our hypothesized $b_0$. We suspect this discrepancy arises due to our prior parameter value $b_0=1$, which is an over-estimate.
Therefore, we compute the difference between that $b_0$ and the posterior predictive density mean $\beta'$ per fitting approach. 
Except for Grid Search ($b_0 - \beta' = -0.002 \pm 0.013$), the approaches boast positive differences (Hyperopt and L-BFGS-B both yield $b_0 - \beta' = 0.183 \pm 0.006$), which imply the prior parameter is larger than the posterior predictive density mean. 
The direction and magnitude of these shifts of the posterior away from the prior suggest the use of our Bayesian approach as a diagnosis tool, which correctly signals that our hypothesis likely over-estimates true $\beta$.

\begin{figure}[!t]
	\begin{center}
		\includegraphics[width=0.62\columnwidth]{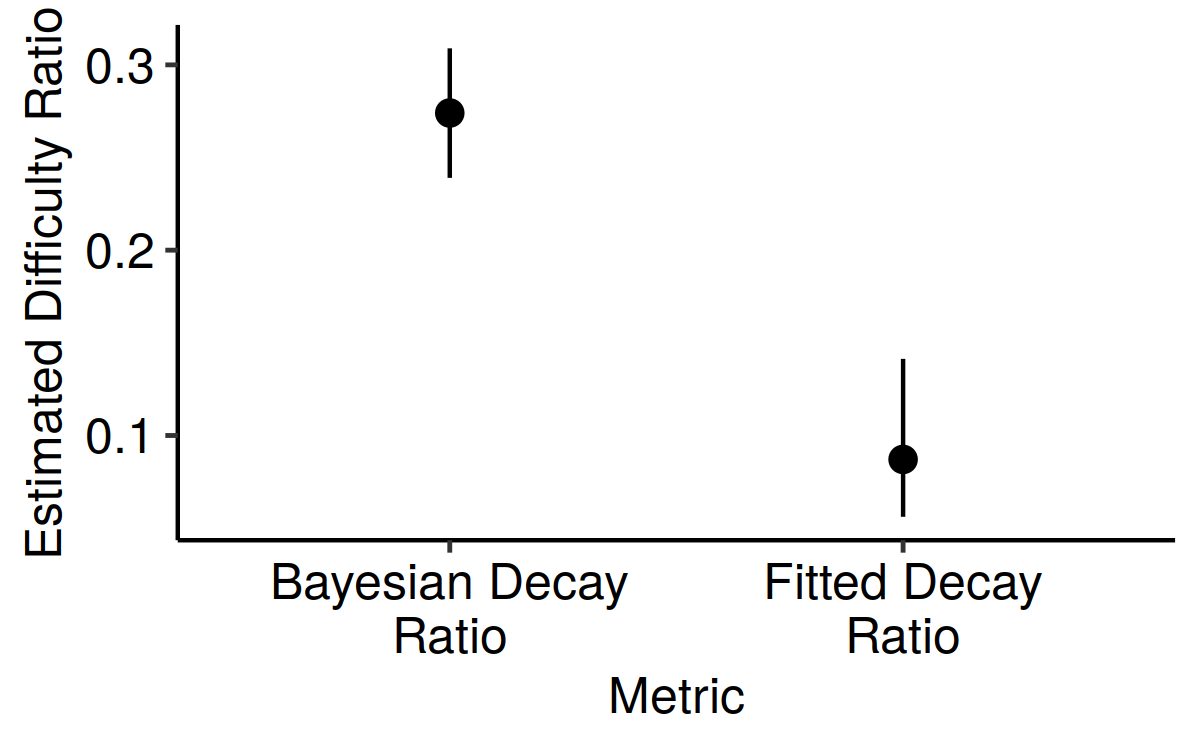}
		\caption{\label{fig:duolingo}
			\textit{In the Duolingo App, Users Learning C-level Words Have Longer Learning Bursts Than Those Learning A-level Words.}
			We fit two Hawkes processes to users with $10$ word learning events on Duolingo: one for users studying ``hard'' (C-level) words, and the other for users learning ``easy'' (A-level) words.
			We posit that the decay value of the C-level process is half as large as the A-level one, and we depict the ratio of the former to the latter. 
			The ratio of fitted values (``Fitted Decay Ratio'') is lower than that of posterior predictive density means (``Bayesian Decay Ratio''), due to a conservative prior parametrization.
			Lower decay values for the same number of events imply longer event bursts, suggesting that C-level words require extended learning effort.
		}
	\end{center}
\end{figure}

\noindent\textbf{Vocabulary Learning Intensity.}
We investigate a scenario proposed as future work by Settles and Meeder's~\cite{settles2016trainable} study of user behavior on the Duolingo language learning app: The authors speculate that vocabulary learning intensity in Duolingo correlates with word difficulty as defined by the CEFR language learning framework~\cite{council2001common}.
We complement the Duolingo data with a dataset of English-language vocabulary and its corresponding CEFR level\footnote{\url{http://www.englishprofile.org/american-english}}, and we build two groups of words: those from the easiest CEFR levels, A1 and A2 (A-level group), and those from the hardest ones, C1 and C2 (C-level group). 
We observe that there are $28$ users with $10$ vocabulary learning events in the C-level. To control for total learning events per user, we randomly sample a set of $28$ users with $10$ learning events in the A-level. 
Increasing the number of events to $11$ or $12$ leads to qualitatively similar results, but decreased statistical power due to smaller sample size.
We repeat this random sampling for a total of $100$ times to Bayesian bootstrap $95\%$ credible intervals. 
We posit that each Duolingo user learning the A-level set of words represents a realization of a univariate Hawkes process, and users learning the C-level words represent realizations of another univariate Hawkes process. 
Mapping the six CEFR levels (A1, A2, B1, B2, C1 and C2) to a scale from $1$ to $6$, we naively assume, for illustration purposes, that the C-levels may be more than twice as hard as A-levels.
If the data reflects that hypothesis, then we expect that a short learning burst suffices for grasping A-level words, in contrast to the C-level words, which may require perhaps more than two times as much effort over time.
We encode this hypothesis in our closed-form Bayesian approach (with L-BFGS-B) as $b_{\text{C-level}}=1$ and $b_{\text{A-level}}=2$, since, after controlling for the total event count, we interpret the former as corresponding to longer periods of higher intensity, when compared with the latter.

In Figure~\ref{fig:duolingo}, we depict the ratio of posterior predictive density means $\beta'_{\text{C-level}} / \beta'_{\text{A-level}}$ (``Bayesian Decay Ratio''), as well the analogue ratio of the means of actual L-BFGS-B estimations for both levels (``Fitted Decay Ratio''), i.e., $\{\hat{\beta}_{\text{C-level}} / \hat{\beta}_{\text{A-level}}\}_{k=1}^{28}$.
Overall, we confirm Settles and Meeder's hypothesis that word difficulty correlates positively with the effort required to learn them: The posterior (and fitted) decay values of the C-level words are lower than those of the A-level words, resulting in more prolonged learning bursts in the former vs. the latter.
Moreover, we underscore that this practical example illustrates the usefulness of our approach as a diagnosis tool: We observe a moderate shift away from the hypothesized $1:2$ difficulty ratio and towards a posterior predictive density mean value slightly below $3:10$. 
This shift is even more severe according to the ``Fitted Decay Ratio'', which is not influenced by a prior.
This finding indicates that the CEFR language levels may not directly translate to numerical scales for quantifying learning progress. 
Further, the relatively small size of the user sample highlights the importance of using Bayesian inference to not only diagnose a-priori estimates, but also to explicitly surface estimation uncertainty due to a small dataset. 
Therefore, although we suggest caution in extrapolating our results, we believe that our observations may contribute to ongoing research on the challenges of quantifying language learning progress~\cite{hulstijn2007shaky}.

\subsection{Addressing Breaks in Stationarity}
We turn our attention to the assumption of stationarity and its effect on fitting the decay parameter of Hawkes processes.
Recall that stationarity implies that the intensity of a Hawkes process is translation-invariant.
In practical applications such as the study of virality of online content~\cite{rizoiu2017expecting} or the growth of online communities~\cite{santos2019self}, exogenous shocks or exponential growth break the stationarity assumption.
With synthetic data and a real-world example, we show how our Bayesian approach allows for assessing and capturing breaks in stationarity caused by exogenous shocks.

\noindent\textbf{Synthetic Data.} 
\label{subsubsec:intractable_bayes}
We start with the same experimental setup as in section~\ref{subsubsec:closed_bayes}, but we introduce two key differences.
First, we assume that there was an underlying change in $\beta$ at some point during the $K=100$ realizations. We set the index of that change to $k^*=50$, but our conclusions also hold for other choices of $k^*$ (such as $k^* \in [30, 70]$). For $k<k^*$, we simulate the Hawkes process as previously, but, for $k\geq k^*$, we increment $\beta$ by $1$ while keeping the other parameters unchanged, and we simulate from that updated process instead. With $\beta_1$ we denote $\beta$ before the change, and with $\beta_2 = \beta_1 + 1$ the $\beta$ afterwards.
Second, we build a Bayesian inference setup to reflect the hypothesis that the $\beta$ value changed at some point in the set of realizations. Such models are termed \textit{changepoint models}. We propose the following intractable setup: A prior with $b_0 = \left\{ \begin{array}{ll} b_1,&k < \kappa\\ b_2,&k \geq \kappa \\ \end{array} \right.$, where we set $b_1\sim Exponential(1)$, $b_2\sim Exponential(0.7)$ and $\kappa \sim U\{1, 100\}$, and an exponentially distributed likelihood. Note that the hypothesis $b_1 > b_2$ contradicts the true second part of simulated realizations. 
As the metrics for this experiment, we first measure the RMSE between the mean (respectively median) of samples from the posterior, $\bar{\beta}_1$ and $\bar{\beta}_2$ (resp. $\kappa / K$), and both true $\beta$ (resp. $k^* / K$). Beyond RMSE, we also assess the estimation accuracy as the relative frequency of a correctly inferred ordering $\bar{\beta}_1 < \bar{\beta}_2$.
As before, we demonstrate that our Bayesian approach is agnostic to the choice of decay fitting methods by employing L-BFGS-B, Grid Search and Hyperopt.

As a side result, we note that L-BFGS-B and Hyperopt again outperform Grid Search with respect to RMSE.
Further, we report mean accuracy values close to $1$ for Hyperopt and L-BFGS-B (both $0.98 \pm 0.014$) and remarkably lower values for Grid Search ($0.65 \pm 0.047$). 
Although we encoded prior parameter values which contradict the data, we still almost always recover the correct relationship $\bar{\beta}_1 < \bar{\beta}_2$ with the L-BFGS-B and Hyperopt methods. 
Qualitatively, this suggests identifying the direction of distributional changes in the decay is feasible.
Quantitatively, we expect this procedure to yield conservative estimates of the magnitude and timing of the change: 
Bayesian inference features an ``inertia'' of a few realizations when updating the posterior after the shock.

\begin{figure}[!t]
	\begin{center}
		\includegraphics[width=0.7\columnwidth]{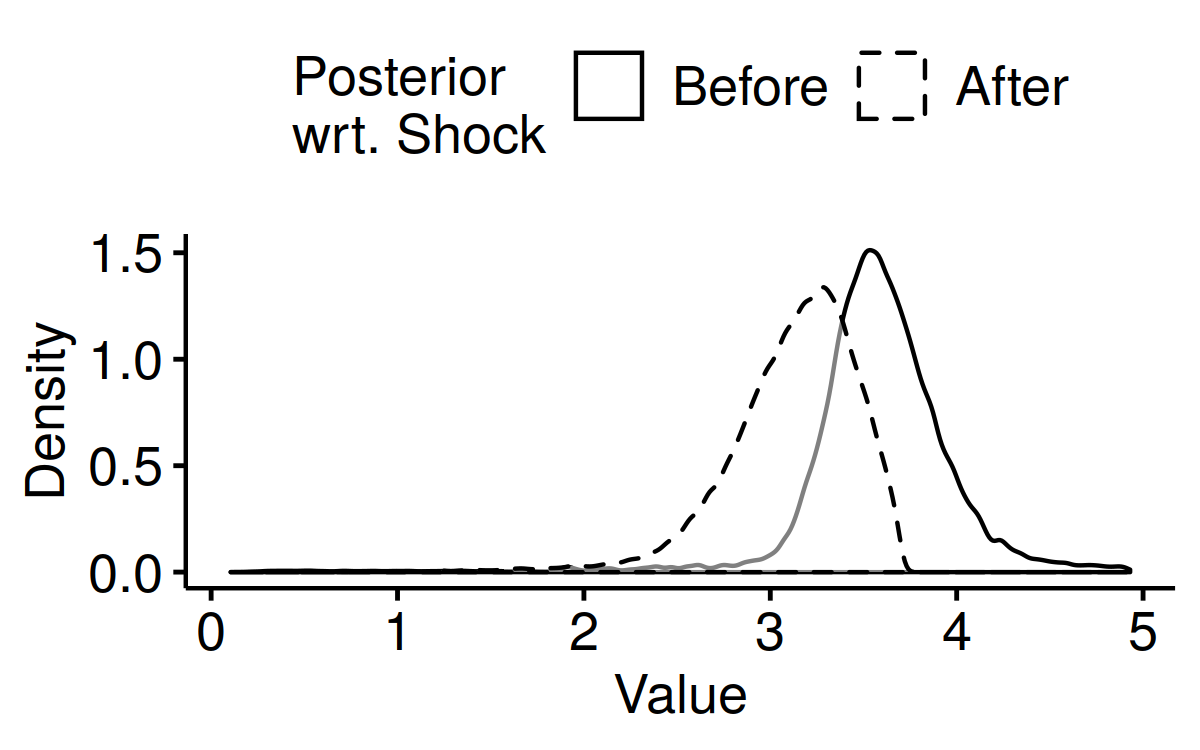}
		\caption{\label{fig:emotions}
			\textit{Tweet Timing Reflects Collective Emotions.}
			We depict the result of fitting an MCMC-based changepoint detection model to users' Tweet timings in the two weeks surrounding the November 2015 Paris attacks: 
			The estimated posterior density for the Hawkes process intensity decay assigns more probability mass to higher decay regions before the shock in comparison to afterwards. This suggests that collective effervescence manifests on Tweet timings, as lower decay values after the shock reflect more sustained bursts of activity. 
		}
	\end{center}
\end{figure}

\noindent\textbf{Strength of Collective Effervescence.}
Our third real-world scenario concerns the manifestation of collective effervescence on Twitter in response to the 13. November 2015 terrorist attacks in Paris, as studied by Garcia and Rim{\'e}~\cite{garcia2019collective}. 
They proposed future work could analyze how tweet timings reflect collective emotions surrounding the attacks.
We address this suggestion by fitting the changepoint Bayesian model with the L-BFGS-B method. 
Specifically, we begin by extracting the timestamps of tweets by users in a two week period centered on the day of the attacks.
We model each user's behavior per week as a realization of a univariate Hawkes process, and we also control for tweeting activity per user: We extract all $205$ users who tweeted between $20$ and $25$ times in the week before and in the week after\footnote{Note that this extraction process results in a total of $410$ realizations, i.e., $205$ realizations before the shock and another $205$ afterwards.}. 
Lowering the activity bounds yields more users each with less events to fit, while increasing those bounds has the opposite effect. However, by setting the activity bounds to different ranges of $5$ tweets (specifically, $10$ to $15$, $15$ to $20$, \ldots, or $45$ to $50$ tweets), we qualitatively observe the same outcomes.
We hypothesize that Twitter users, who partake in the collective emotion as a reaction to the shock, feature a sustained burst of activity. We expect such a burst of activity to translate into a decrease of the decay value after the shock. To numerically capture this hypothesis, we simply set $b_1 = 1.5, b_2=1$.
However, repeating this experiment with an opposing hypothesis (e.g., $b_1 = 1, b_2=1.5$), again leads to the same results.

Figure~\ref{fig:emotions} depicts the density of the distribution of the inferred decay posterior before and after the shock. 
As expected, we confirm the decay value goes down in the week after the attacks, suggesting more sustained bursts of Tweeting activity in response to the attacks.
This, in turn, supports the hypothesis that Garcia and Rim{\'e} advanced: Reaction timings, in the form of longer bursts of tweets afforded by a $15\%$ lower mean posterior decay after the shock, reflect this collective emotion. 
We note that this is a conservative estimate of the decrease in the parameter, since activity levels quickly revert back to a baseline within the week after the attacks themselves, as Garcia and Rim{\'e} report. 
Further, this changepoint detection approach also over-estimates the time of the change at realization number $235$, i.e., $7.1\%$ later than the first Hawkes process realization after the shock, corresponding to realization number $206$. 

\section{Discussion}
\label{sec:relwork}
In this section, we first reflect on practical aspects of our Bayesian approach vs. a frequentist alternative.
Then, we relate our work to literature on Hawkes process applications and theory. 

\noindent \textbf{Comparison With Bootstrap.}
Bayesian and frequentist inference are two intensively discussed alternative statistical schools for estimating parameters. In recent years, Bayesian approaches received increasing appreciation by the web and machine learning communities (e.g.~\cite{barber2012bayesian,yang2012bayesian,singer2015hyptrails}). Beyond that body of work and theoretical considerations (cf. also Sec.~\ref{sec:approach}), we outline practical differences between our Bayesian approach and a bootstrap-based frequentist alternative. 
Quantifying the uncertainty of decay estimates with bootstrap amounts to deriving confidence (rather than credible) intervals for the decay from the empirical bootstrap distribution.
Diagnosing mis-estimates and misaligned hypotheses with the bootstrapped decay distribution is not an integral part of the frequentist inference procedure, on the contrary to our Bayesian approach, which integrates the hypothesis in the inference procedure. However, explicitly formulating one- or two-sided statistical hypotheses to be tested against the bootstrap distribution is just as viable. 
In the case of addressing breaks of stationarity, the bootstrap-based approach features remarkably higher complexity than our proposed Bayesian approach. While we infer all three parameters capturing the stationarity break simultaneously, the straightforward bootstrap-based alternative would require alternatively computing the distribution of one parameter while fixing the other two (as the inferred distribution is intractable). Hence, we argue that our Bayesian approach may also be the more natural choice for intractable inference.
Nevertheless, we stress that the frequentist alternative we outlined is a viable alternative to our Bayesian approach.

\noindent \textbf{Related Work.}
One of the first fields to leverage the seminal work by Hawkes~\cite{hawkes1971spectra} includes seismology~\cite{ogata1982application,daley2003introduction}.
Since then, Hawkes process theory and practice emerged in the realm of finance~\cite{bacry2015hawkes,bacry2016mean,trouleau2019learning}, as well as, more recently, in modeling user activity online~\cite{zhou2013learning,zhou2013learningt,upadhyay2017uncovering,tabibian2017distilling,rizoiu2017expecting,kurashima2018modeling,tabibian2019enhancing,santos2019self,junuthula2019block,hatt2020early}.
More specifically, the latter body of work extended Hawkes processes to predict diffusion and popularity dynamics of online media~\cite{zhou2013learningt,zhou2013learning,rizoiu2017expecting}, model online learning~\cite{upadhyay2017uncovering,tabibian2019enhancing},
capture the spread of misinformation~\cite{tabibian2017distilling}, and understand user behavior in online communities~\cite{santos2019self,junuthula2019block}, in online markets~\cite{hatt2020early} and in the context of the offline world~\cite{kurashima2018modeling}. As all of those previous references interpreted the parameter values of Hawkes processes (and variations thereof), they may benefit from our study of the decay parameter, especially as we uncover its properties and assess and mitigate estimation issues with Bayesian inference.

Perhaps closest to our work is Bacry et al.'s~\cite{bacry2016mean} study of mean field inference of Hawkes process values. In particular, those authors inspected the effect of varying the decay parameter across a range of values: With increasing decay, fitted self- and cross-excitations decrease while baseline intensity increases. 
We go beyond their study by deepening our understanding of the (noisy) properties of the Hawkes log-likelihood as a function of the decay. 
Methodologically, our Bayesian approach relates to Hosseini et al.'s~\cite{hosseini2016hnp3}. Those authors infer the decay parameter by assuming a Gamma prior and computing the mean of samples from the posterior (as part of a larger inference problem). In our work, we instead focus on the Bayesian approach as a means to quantify estimation uncertainty.
Further, as our Bayesian changepoint model captures breaks in stationarity, we simplify previous work~\cite{rizoiu2017expecting,santos2019self} which relies on additional assumptions, such as estimating stationarity via the time series of event counts. 
Finally, our work complements recent efforts~\cite{trouleau2019learning} to learn Hawkes processes from small data.

\section{Conclusion}
In this work, we studied the problem of fitting the decay parameter of Hawkes processes with exponential kernels. 
The inherent difficulties we found in accurately estimating the decay value, regardless of the fitting method, relate to the noisy, non-convex shape of the Hawkes log-likelihood as a function of the decay. 
Further, we identified problems in quantifying uncertainty and diagnosing fitted decay values, as well as in addressing breaks of the stationarity assumption. As a solution, we proposed a parsimonious Bayesian approach. 
We believe our extensive evaluation of that approach across a range of synthetic and real-world examples demonstrates its broad practical use. 

Optimization techniques such as constructing convex envelopes or disciplined convex-concave programming may, in the future, help to optimize the Hawkes process likelihood as a function of the decay.
We also believe exploring the potential of the vast Bayesian statistics toolbox for learning more from fitted (decay) parameter values is promising future work.

\section*{Acknowledgments}
We thank David Garcia for giving access to the extended Twitter data.
We also thank Roman Kern and the anonymous reviewers for their valuable feedback on the manuscript.
Tiago Santos was a recipient of a DOC Fellowship of the Austrian Academy of Sciences at the Institute of Interactive Systems and Data Science of the Graz University of Technology.

\balance
\bibliographystyle{IEEEtran}
\bibliography{bib}

\end{document}